\documentclass{article}

\usepackage{PRIMEarxiv}

\usepackage[utf8]{inputenc} 
\usepackage[T1]{fontenc}    
\usepackage{hyperref}       
\usepackage{url}            
\usepackage{booktabs}       
\usepackage{amsfonts}       
\usepackage{nicefrac}       
\usepackage{microtype}      
\usepackage{lipsum}
\usepackage{fancyhdr}       
\usepackage{graphicx}       
\graphicspath{{media/}}     
\usepackage{color}
\usepackage{amsmath}
\usepackage{arydshln}
\title{DetectGPT-SC: Improving Detection of Text Generated by Large Language Models through Self-Consistency with Masked Predictions
}

\author{
  Rongsheng Wang \\
  Macao Polytechnic University \\
  \texttt{p2213046@mpu.edu.mo} \\
   \And
  Qi Li \\
  Iowa State University \\
  \texttt{qli@iastate.edu} \\
   \And
  Sihong Xie\thanks{Corresponding author} \\
  HKUST (GZ) \\
  \texttt{sihongxie@hkust-gz.edu.cn} \\
}

\begin{document}
\maketitle

\begin{abstract}
General large language models (LLMs) such as ChatGPT have shown remarkable success, but it has also raised concerns among people about the misuse of AI-generated texts. Therefore, an important question is how to detect whether the texts are generated by ChatGPT or by humans. Existing detectors are built on the assumption that there is a distribution gap between human-generated and AI-generated texts. These gaps are typically identified using statistical information or classifiers. In contrast to prior research methods, we find that large language models such as ChatGPT exhibit strong self-consistency in text generation and continuation. Self-consistency capitalizes on the intuition that AI-generated texts can still be reasoned with by large language models using the same logical reasoning when portions of the texts are masked, which differs from human-generated texts. Using this observation, we subsequently proposed a new method for AI-generated texts detection based on self-consistency with masked predictions to determine whether a text is generated by LLMs. This method, which we call DetectGPT-SC. We conducted a series of experiments to evaluate the performance of DetectGPT-SC. In these experiments, we employed various mask scheme, zero-shot, and simple prompt for completing masked texts and self-consistency predictions. The results indicate that DetectGPT-SC outperforms the current state-of-the-art across different tasks.
\end{abstract}


\section{Introduction}
Large language models (LLMs), such as ChatGPT, represent a significant milestone in the field of natural language processing (NLP). Large models have been pre-trained on extensive text corpora, enabling them to generate texts that is contextually relevant and fluent. They have greatly advanced various NLP tasks, including text generation, question-answering, and text classification. Recent research findings suggest that responses generated by large language models, while often appearing highly persuasive, frequently contain inaccuracies. Nevertheless, in specific contexts, the lucidity of the texts produced by these models can prove enticing, particularly within the domains of student paper composition and news reporting. The application of these extensive language models presents multifaceted challenges, not only rendering equitable student assessment more complex but also impeding student learning while amplifying the presence of persuasive yet erroneous news articles. Unfortunately, when classifying AI-generated and human-generated texts, humans perform only slightly better than random guessing~\cite{1-2}, leading researchers to consider automated detection that can identify signals that are difficult for humans to detect.

\begin{figure}[h]
  \centering
  \includegraphics[width=\linewidth, trim=230 95 230 60, clip]{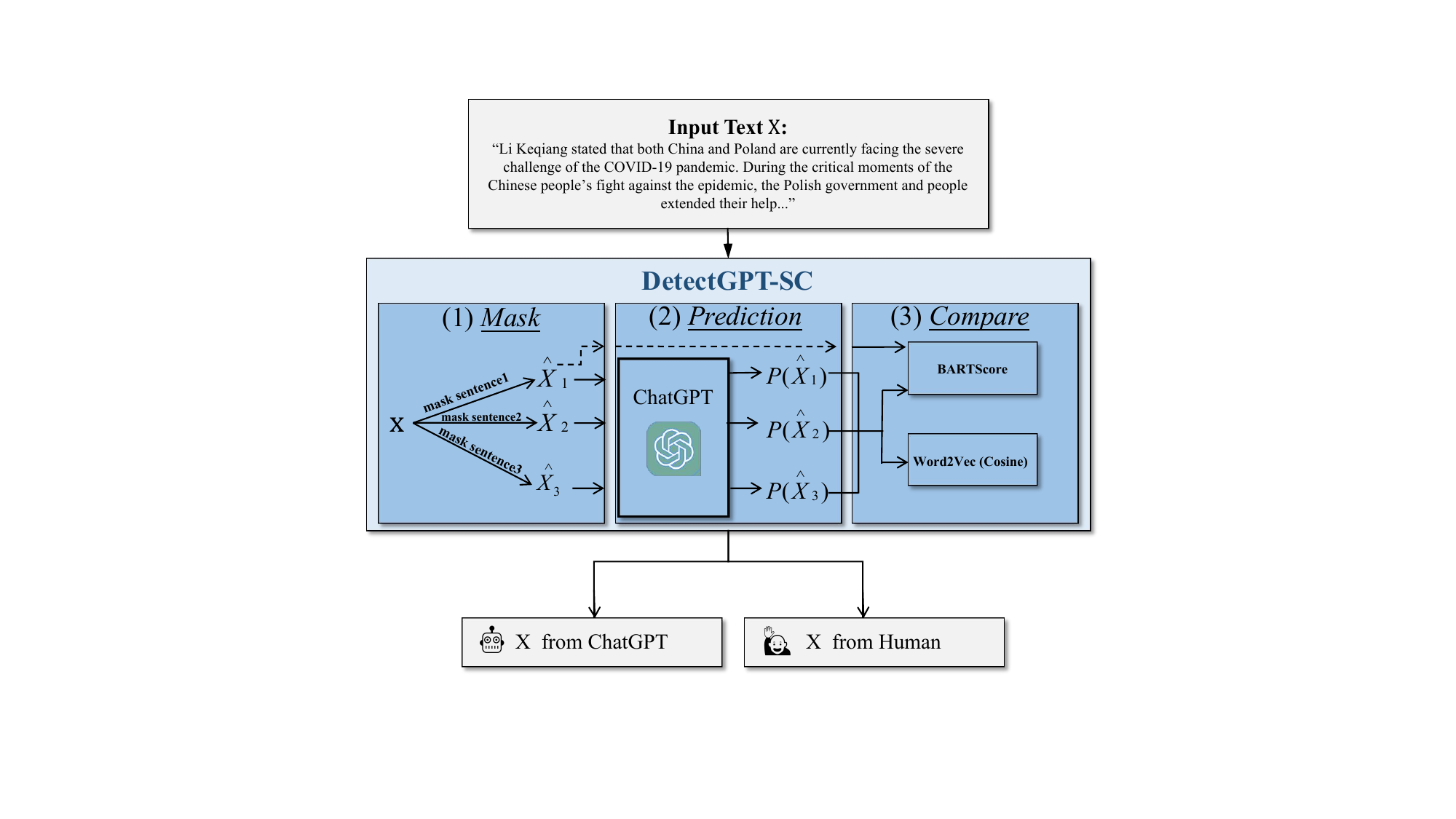}
  \caption{Self-consistency with masked predictions consists of three steps: (1) Masking the input text $X$; (2) Applying self-consistency with masked predictions to the masked text; (3) Calculating the similarity between the predicted masked text and the original masked text.}
  \label{fig1}
\end{figure}

The foundation of current text detection is based on an assumption that there is a distributional difference between AI-generated texts and human-generated texts. These differences are typically ascertained by employing statistical information or training classifiers~\cite{2-13}. Recent research has delved deeper into understanding the unique capabilities of large language models, such as self-consistency as introduced by Wang et al.~\cite{2-12}. Wang et al.~\cite{2-12} introduced a new decoding strategy called self-consistency. Self-consistency addresses complex reasoning tasks by extracting a variety of reasoning paths and selecting the most consistent answer. We have observed that self-consistency unveils the intrinsic advanced logical features and expressiveness within LLMs. Specifically, when portions of a text are masked, LLMs exhibit robust logical reasoning capabilities to accurately infer the obscured content. If the texts inferred by the LLMs aligns highly in structure and semantics with the original obscured content, it provides a credible basis to attribute the texts to being generated by LLMs. The method we propose is shown in Figure~\ref{fig1}.

The contributions of this paper are summarized as follows:
\begin{itemize}
     \item \textbf{Dataset development}: To promote research related to LLMs, especially studies comparing humans and LLMs, we gathered nearly 4,000 news texts from Yangshi News and ChatGPT, naming it the ChatGPT and Yangshi News (CYN) dataset. The CYN dataset serves as a valuable resource for analyzing the linguistic and stylistic characteristics of both humans and ChatGPT while also demonstrating the limitations of domain-specific classifiers.
    \item \textbf{Accurate domain invariant detection}: We introduce a strategy of implementing self-consistency with masked predictions in LLMs to enhance the accuracy of text detection. Our method overcomes the constraints of training separate classifiers, amassing extensive datasets of real or generated texts, or explicitly adding watermarks to the generated content. It can be realized by simply interacting with LLMs using prompts. In text detection across different domains, DetectGPT-SC achieved an accuracy of 91.1\% on HC3 and 93.3\% on CYN, demonstrating strong detection and generalization capabilities.
    \item \textbf{Comprehensive empirical evaluation}: We conducted extensive empirical evaluations using the most advanced text detectors ChatGPT Detector~\cite{2-2}, GLTR~\cite{1-2}, PPL~\cite{2-2}, Roberta-base-openai-detector~\cite{1-3}, and DetectGPT~\cite{2-3} on CYN dataset and HC3 dataset.
\end{itemize}

\section{Related Work}
Recent research has shown promising results in the development of detection methods. The existing detectors are built on the assumption that there is a distributional difference between human-generated texts and AI-generated texts. These differences are typically identified by training classifiers or using statistical information.

\textbf{Classifier-based detectors.} Classifier-based detectors are commonly used in natural language processing detection paradigms, especially in fake news and misinformation detection~\cite{2-1}. Guo et al.~\cite{2-2} proposed the ChatGPT Detector, where they initially constructed a dataset consisting of ChatGPT conversations with human questions and answers, and trained a text detection classifier based on this dataset. The use of these methods requires substantial data collection and incurs the cost of training these classifier models.

\textbf{Statistical-based detectors.} Statistical-based detectors utilize statistical metrics such as entropy, perplexity, and $n$-gram frequency to differentiate between human-generated and AI-generated texts~\cite{1-2}. 

Some other work is based on watermark-based detectors. In previous research, watermarks have been applied in the field of image processing and computer vision to protect copyrighted content and prevent intellectual property theft~\cite{2-4}. Recently, with the emergence of ChatGPT, the work by Kirchenbauer et al.~\cite{2-5} demonstrated how to incorporate a watermark using only the logarithmic credentials of each step to mark AI-generated texts. While watermark-based detectors are an intriguing area of research, adding watermarks may affect the readability of the texts, and the removal of watermarks is also a challenge we need to address.

However, with the emergence of ChatGPT, an innovative statistical detection method called DetectGPT~\cite{2-3} has been developed. Its principle is that text generated by the model typically resides in the negative curvature region of the model's log probability. DetectGPT~\cite{2-3} generates and compares multiple variants of model-generated texts to determine whether the texts are machine-generated based on the log probabilities of the original texts and these variants. DetectGPT~\cite{2-3} outperforms the vast majority of existing zero-shot methods in terms of model sample detection, achieving very high AUC. It is based on this concept that DetectGPT-SC was proposed.

\section{Methodology}

\textbf{Predefined definition} $X$ is the original input text. ${\hat{X}}$ is the sentence masked within $X$, which can also be referred to as "<mask>". $P({\hat{X}})$ is the mask content predicted by ChatGPT.

DetectGPT-SC is based on the hypothesis that AI-generated texts can still undergo reasoning by large language models using the same logical reasoning process even when certain portions of the texts are masked, which distinguishes it from texts generated by humans. DetectGPT-SC is shown in Figure~\ref{fig1}, by masking the input text $X$ multiple times, we obtain ${\hat{X}_{1}}$, ${\hat{X}_{2}}$, and ${\hat{X}_{3}}$, which represent the masked content in $X$. Simultaneously, we replace the corresponding positions in the input text $X$ with the "<mask>" token. Then, we utilize ChatGPT to generate completions for the masked ${\hat{X}_{1}}$, ${\hat{X}_{2}}$, and ${\hat{X}_{3}}$, resulting in new inferred results: $P({\hat{X}_{1}})$, $P({\hat{X}_{2}})$, and $P({\hat{X}_{3}})$. Next, we employ cosine similarity and Word2Vec to compute the similarity between $P({\hat{X}_{i}}), i=1,2,3$ and ${\hat{X}_{i}}, i=1,2,3$. Word2Vec is a word embedding model that represents words as vectors and captures the semantic relationships between them. By calculating the cosine similarity between $P({\hat{X}_{i}}), i=1,2,3$ and ${\hat{X}_{i}}, i=1,2,3$ in the vector space, we can evaluate their semantic similarity. The cosine similarity is shown as follows:

\begin{equation}
\text{BARTSCORE} = \sum_{t=1}^{m} \omega_{t} \log p\left(\mathbf{y}{t} \mid \mathbf{y}{<t}, \mathbf{x}, \theta\right) \tag{1}
\end{equation}

Finally, we utilize BARTScore~\cite{4-2} to calculate the similarity between $P({\hat{X}})$ and $X$, which enables us to assess the similarity between the generated texts and the original text. BARTScore compares the similarity between two text sequences and provides a comprehensive similarity score. It evaluates the overall similarity between the generated texts and the original text, including aspects such as syntactic structure and contextual coherence. The BARTScore is shown as follows:


\begin{equation}
\text{Cosine Similarity} = \frac{P({\hat{X}}) \cdot X}{\left|P({\hat{X}})\right| \cdot \left|X\right|} \tag{2}
\end{equation}

Through these methods, we can reason with AI-generated texts and measure the similarity between the generated texts and the original text. Table~\ref{tab1} displays examples of self-consistency with masked predictions.

\begin{table*}
  \caption{Examples of self-consistency with masked predictions}
  \label{tab1}
  \begin{tabular}{p{\textwidth}}
    \toprule
    \textbf{Text} \\
    On the evening of April 16, Premier Li Keqiang of the State Council received a scheduled call from Polish Prime Minister Morawiecki. Li Keqiang stated that both China and Poland are currently facing the severe challenge of the COVID-19 pandemic. During the critical moments of the Chinese people’s fight against the epidemic, the... (Here are some omitted contents.)\\
    \midrule
    \textbf{Mask} \\
    On the evening of April 16, Premier Li Keqiang of the State Council received a scheduled call from Polish Prime Minister Morawiecki. \colorbox{yellow}{<mask1>}. During the critical moments of the Chinese people’s fight against the epidemic, the Polish government and people extended their help. \colorbox{yellow}{<mask2>}. \colorbox{yellow}{<mask3>}.\\
    \midrule
    \textbf{Prompt} \\
    \colorbox{green}{Given the following sentence, please complete <mask>}: On the evening of April 16, Premier Li Keqiang of the State Council received a scheduled call from Polish Prime Minister Morawiecki. <mask1>. During the critical moments of the Chinese people’s fight against the epidemic, the Polish government and people extended their help. <mask2>. <mask3>.\\
    \midrule
    \textbf{Output} \\
    On the evening of April 16, Premier Li Keqiang of the State Council received a scheduled call from Polish Prime Minister Morawiecki. They discussed the ongoing cooperation between the two countries and reaffirmed their commitment to strengthening diplomatic ties. During the critical moments of the Chinese people's fight against the epidemic, the Polish government and people extended their help. This solidarity between nations serves as a testament to the importance of international collaboration in times of crisis. It highlights the global community's resilience in the face of adversity and underscores the significance of working together to overcome common challenges.\\ \\
    \hdashline
    \\
    \colorbox{yellow}{<mask1>}: They discussed the ongoing cooperation between the two countries and reaffirmed their commitment to strengthening diplomatic ties. \\
    \colorbox{yellow}{<mask2>}: This solidarity between nations serves as a testament to the importance of international collaboration in times of crisis. \\
    \colorbox{yellow}{<mask3>}: It highlights the global community's resilience in the face of adversity and underscores the significance of working together to overcome common challenges. \\
    \bottomrule
  \end{tabular}
\end{table*}

\section{Experiments}

In this section, we provide detailed empirical evidence to support our claim on the detectability of this work, including benchmarks, evaluation metrics, text detectors, mask scheme and sampling scheme.

\textbf{Benchmarks:} We have created a dataset called ChatGPT and Yangshi News (CYN) dataset, which comprises nearly 4K samples from both Yangshi News texts and ChatGPT-generated texts. The human-generated texts were collected from Yangshi News, and we sampled 1,700 samples. On the other hand, the ChatGPT-generated texts were sampled using prompts in ChatGPT (gpt-3.5-turbo-16k). During the generation process, we used temperature is 0.7. The temperature is used to control the randomness when generating texts. Temperature is a floating-point number between 0 and 1, and it influences the diversity and creativity of the model's output. The Human ChatGPT Comparison Corpus (HC3) dataset~\cite{2-2} comprises responses from both ChatGPT and human volunteers to identical questions, spanning various domains such as open-domain, computer science, finance, medicine, law, and psychology. The dataset is shown in Table~\ref{tab2}.

\begin{table}
  \begin{center}
    \caption{Evaluation Dataset and Sample Size}
    \label{tab2}
    \begin{tabular}{ccc}
      \toprule
      Dataset & Human-generated & ChatGPT-generated \\
      \midrule
      CYN & 1,700 & 1,918 \\
      HC3~\cite{2-2} & 8,136 & 10,455 \\
      \midrule
      ALL & 9,836 & 12,373 \\
      \bottomrule
    \end{tabular}
  \end{center}
\end{table}

\textbf{Metrics:} We use accuracy (ACC) to measure the results of text classification.

\textbf{Detectors and prompt:} We examined several publicly text detectors.

\begin{itemize}
\item \textbf{Classifier-based detectors}: ChatGPT Detector~\cite{2-2} and Roberta-base-openai-detector~\cite{1-3}.
\item \textbf{Statistical-based detectors}: GLTR~\cite{1-2}, PPL~\cite{2-2} and DetectGPT~\cite{2-3}.
\end{itemize}

We conducted all experiments in the zero-shot setting without training or fine-tuning the language models. For a fair comparison, we used the same prompt: "Given the following sentence, please complete <mask>:".

\textbf{Mask scheme:} Before sampling in ChatGPT, we selectively masked certain parts of the text (left sentence, center sentence, right sentence, random sentence) to encourage the model to infer missing information. Specifically, we used special '<mask>' tokens to replace words or even entire sentences. When using the left sentence mask, center sentence mask, random sentence mask, or right sentence mask, if the texts exceeded 400 words, we masked three sentences, this is referred to as "3-mask". If the texts were less than 400 words, we masked one sentence, this is referred to as "1-mask". In Section~\ref{sec:Ablation}, we evaluated the effectiveness of our masking scheme.

\textbf{Sampling scheme:} To sample masked text content in ChatGPT, we adopted a similar method to that proposed by Holtzman et al.~\cite{4-1} in open-text generation. Specifically, for ChatGPT, we used temperature sampling. Temperature is 0.7 (or 0.1) and max\_tokens is 2048. In Section~\ref{sec:Ablation}, we provided an ablation study to demonstrate the robustness of the sampling strategies and parameters.

\subsection{Results}

Following the aforementioned experimental setup, we conducted comprehensive empirical studies. Table~\ref{tab3} displays the accuracy.

\begin{table}
  \centering
  \caption{Accuracy (\%) of different detectors on each dataset}
  \label{tab3}
  \begin{tabular}{lcccc}
    \toprule 
    & \multicolumn{3}{c}{CYN} & HC3\\ 
    \cmidrule(lr){2-4} \cmidrule(lr){5-5} 
    Detectors & $ACC_{1mask}$ & $ACC_{3mask}$ & ACC & ACC\\ 
    \midrule
    ChatGPT Detector~\cite{2-2} & 82.6\% & 91.0\% & 88.3\% & \textbf{98.8\%}\\
    GLTR~\cite{1-2} & 71.0\% & 76.7\% & 75.1\% & 84.1\%\\
    PPL~\cite{2-2} & 48.5\% & 46.4\% & 47.0\% & 44.9\%\\
    Roberta-openai~\cite{1-3} & 50.6\% & 53.4\% & 53.1\% & 49.3\%\\
    DetectGPT~\cite{2-3} & 77.4\% & 81.4\% & 79.1\% & 87.5\%\\
    \midrule
    \textbf{DetectGPT-SC (ours)} & \textbf{88.1\%} & \textbf{94.4\%} & \textbf{93.3\%} & \textbf{91.1\%}\\
  \bottomrule
  \end{tabular}
\end{table}

\begin{figure*}
  \includegraphics[width=\textwidth]{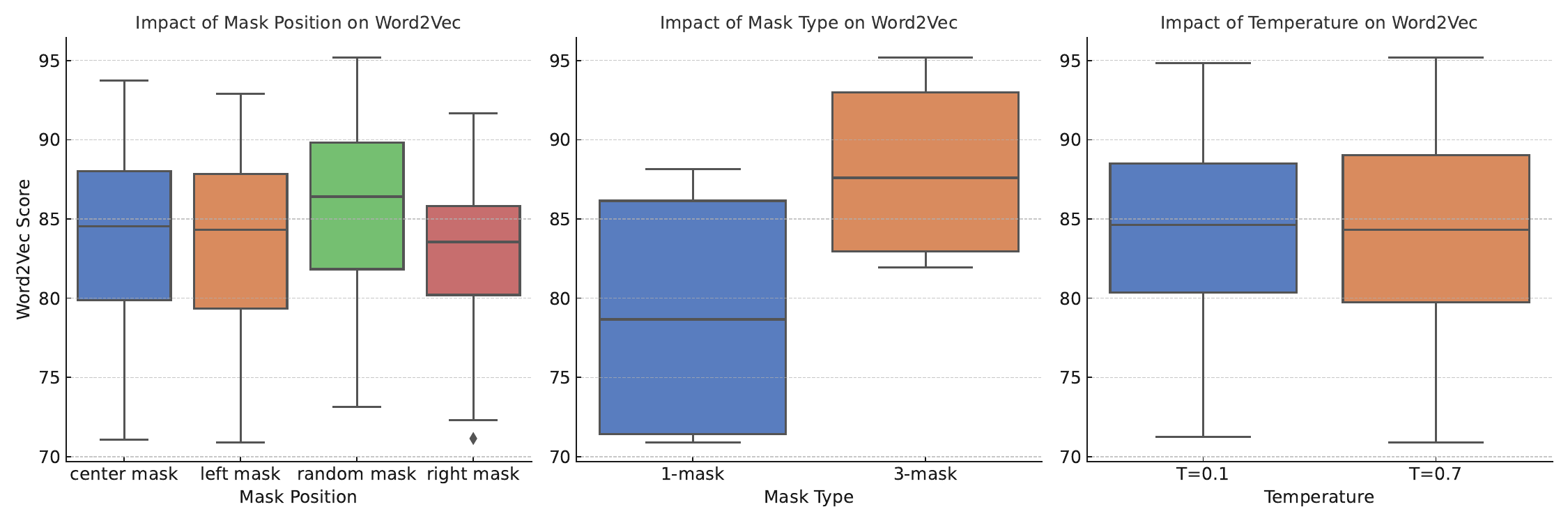}
  \caption{Comparison of the impact of mask type, mask position, and temperature on Word2Vec.}
  \label{fig2}
\end{figure*}

\begin{figure*}
  \includegraphics[width=\textwidth]{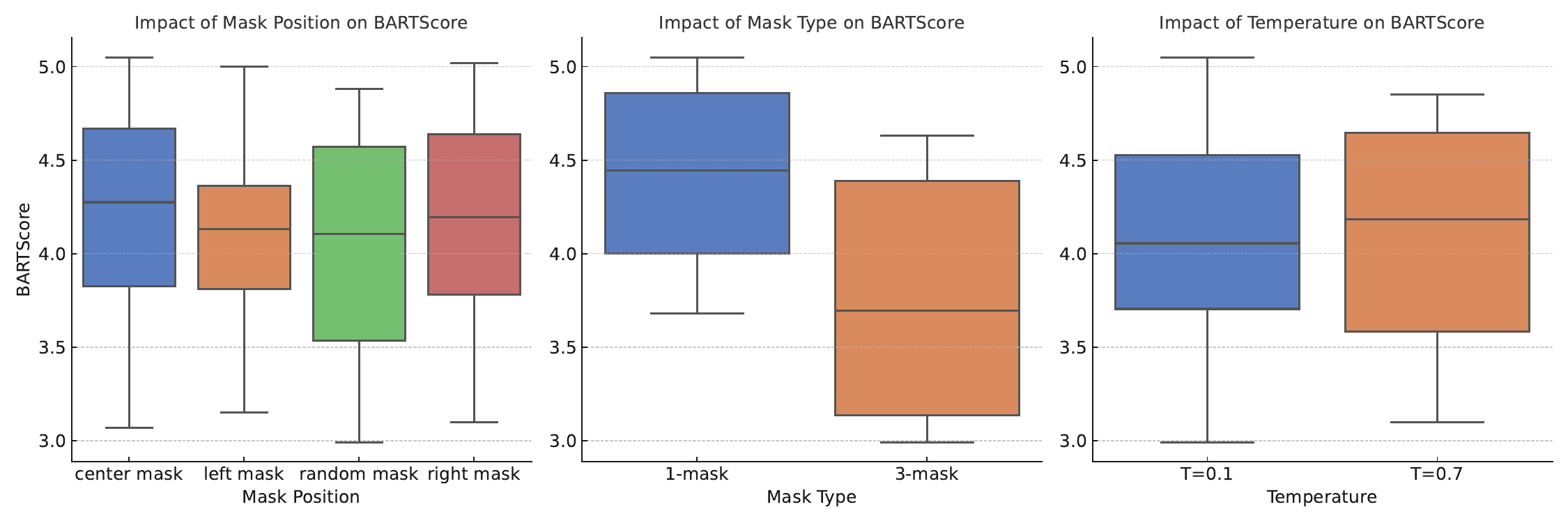}
  \caption{Comparison of the impact of mask type, mask position, and temperature on BARTScore.}
  \label{fig3}
\end{figure*}

\textbf{DetectGPT-SC is effective and generalizes across all benchmarks.} In two benchmarks, DetectGPT-SC demonstrated excellent detection performance. For PPL and Roberta-openai, the accuracy of text detection on CYN and HC3 was around 50\%. DetectGPT-SC achieved the highest score in terms of accuracy. However, we observed that on the HC3 dataset, ChatGPT Detector outperformed DetectGPT-SC. ChatGPT Detector was specifically trained on the HC3 dataset, which allowed it to adapt to the dataset's specific characteristics, patterns, and nuances. 

\textbf{Longer masked text content can achieve higher accuracy.} For the majority of detectors, their performance on $ACC_3mask$ is better than on $ACC_1mask$ in CYN benchmark. This may be due to the fact that longer text provides a richer context, helping the model capture more semantic and syntactic patterns, thereby accurately identifying text regions. The abundance of contextual information assists the model in making more informed decisions and reducing ambiguity in text detection.

\subsection{Ablation Study}\label{sec:Ablation}

In this section, we took into account two key factors that have a significant impact on text detection: the length of masked text content (Mask Type in figure) and the position of the mask (Mask Position in figure). Additionally, we conducted in-depth research on the effect of temperature sampling in the ChatGPT model on its performance and stability. Here it is important to note that BARTScore is a negative value. In order to better visualize it in the figure, we take the absolute value, so a smaller BARTScore indicates better performance.

\textbf{The impact of mask position on the score.} In Figure~\ref{fig2} and Figure~\ref{fig3}, the influence of Mask Position on the score is particularly noticeable. In the scoring systems of Word2Vec and BARTScore, scores derived using random Mask Positions are relatively higher. In other words, random mask positions may facilitate the model to capture and learn the intrinsic characteristics and structures of text data more deeply and comprehensively.

\textbf{The impact of mask type on the score.} In Figure~\ref{fig2} and Figure~\ref{fig3}, the performance of 3-mask type is significantly better than that of 1-mask type. When using 1-mask, the model overly relies on limited surrounding context for predictions. This results in limited generalization ability of the model. However, when using 3-mask, the model relies on a deeper understanding of the surrounding context to correctly predict the masked tokens. This requirement for more precise predictions encourages the model to learn more meaningful and relevant representations.

\textbf{The impact of ChatGPT's temperature on the score.} In Figure~\ref{fig2} and Figure~\ref{fig3}, the performance of a temperature of 0.1 is slightly better than 0.7, but this difference is not significant. ChatGPT's temperature controls the randomness or diversity of the generated texts. A higher temperature (such as 0.7) encourages more randomness in the generated texts, while a lower temperature (such as 0.1) makes the generated texts more deterministic or focused. In scoring systems, the primary evaluation is focused on the quality and coherence of the generated texts. These metrics emphasize the overall quality of the generated output rather than the specific level of randomness or diversity.

\section{Conclusion}
In this paper, we have successfully introduced and validated DetectGPT-SC, which distinguishes itself significantly from existing literature. Our work proposes an innovative text detection method based on the self-consistency and masked prediction of text generated by large language models. Experimental results demonstrate that DetectGPT-SC achieves higher accuracy and efficiency compared to current state-of-the-art text detector. It does not rely on complex classifiers or extensive training data. We hope that our work will provide a valuable starting point for future research and inspire further exploration of the potential of self-consistency in language models for text detection.



\end{document}